# Using large language models to produce literature reviews: Usages and systematic biases of microphysics parametrizations in 2699 publications

**Tianhang Zhang[1], Shengnan Fu[1,2], David M. Schultz[1,3] and Zhonghua Zheng[1,3]**

[1]Centre for Atmospheric Science, Department of Earth and Environmental Sciences, University of Manchester, Manchester, United Kingdom.

[2]School of Hydraulic Engineering, Dalian University of Technology, Dalian 116024, China.

[3]Centre for Crisis Studies and Mitigation, University of Manchester, Manchester, United Kingdom.

Corresponding author: Shengnan Fu (lalaxiaoer@mail.dlut.edu.cn)

**Key Points:**

- A large language model is applied to a literature review of microphysics parameterizations.
- The use of one-moment parameterizations was more common before 2020, whereas two-moment schemes became more prevalent afterward.
- Seven out of nine common microphysics parameterizations tended to overestimate precipitation.

## Abstract

Large language models afford opportunities for using computers for intensive tasks, realizing research opportunities that have not been considered before. One such opportunity could be a systematic interrogation of the scientific literature. Here, we show how a large language model can be used to construct a literature review of 2699 publications associated with microphysics parametrizations in the Weather and Research Forecasting (WRF) model, with the goal of learning how they were used and their systematic biases, when simulating precipitation. The database was constructed of publications identified from Web of Science and Scopus searches. The large language model GPT-4 Turbo was used to extract information about model configurations and performance from the text of 2699 publications. Our results reveal the landscape of how nine of the most popular microphysics parameterizations have been used around the world: Lin, Ferrier, WRF Single-Moment, Goddard Cumulus Ensemble, Morrison, Thompson, and WRF Double-Moment. More studies used one-moment parameterizations before 2020 and two-moment parameterizations after 2020. Seven out of nine parameterizations tended to overestimate precipitation. However, systematic biases of parameterizations differed in various regions. Except simulations using the Lin, Ferrier, and Goddard parameterizations that tended to underestimate precipitation over almost all locations, the remaining six parameterizations tended to overestimate, particularly over China, southeast Asia, western United States, and central Africa. This method could be used by other researchers to help understand how the increasingly massive body of scientific literature can be harnessed through the power of artificial intelligence to solve their research problems.

## Plain Language Summary

Microphysics parameterization is a method used in mesoscale models to simulate unresolved microphysical processes and hydrometeor populations at sub-grid scale in clouds and precipitation. However, to our knowledge, no systematic literature review, meta-analysis, or other large study has understood the usage and performance of different microphysical parametrizations. This study used a large language model to analyze 2699 scientific publications associated with precipitation simulations. China (particularly southeast China and the Sichuan Basin) and the central United States have the most scientific publications on precipitation. The most popular parameterization was WSM6 before 2019, but Thompson after 2020. Seven out of nine parametrizations tended to overestimate precipitation. However, their specific performance varies depending on the region. Our results provide a guide for the choice of microphysics parameterizations and promote the development and accuracy of rainfall forecasts and simulations. Moreover, this study demonstrates a novel approach to solving scientific problems through large-scale analysis of existing scientific publications.

# 1 Introduction

Clouds and precipitation in mesoscale models are represented by microphysics parameterizations that calculate unresolved microphysical processes and hydrometeor populations at sub-grid scale. Because direct observational evidence and our knowledge of these processes is often limited (e.g., Grabowski et al., 2019; Morrison et al., 2020), such parameterizations are often constructed using approximate, hypothesized, and empirical formulas. Early microphysics parameterizations were relatively simple. For example, the single-moment Kessler scheme (Kessler, 1969) predicts only the mixing ratio of cloud and rain drops, while prescribing particle number concentrations based on empirical formulas. As computer power has increased, more sophisticated parameterizations have been developed and become operational. For example, double-moment schemes (e.g., Hong et al., 2010; Lim & Hong, 2010; Milbrandt & Yau, 2005; Morrison & Grabowski, 2008; Thompson et al., 2008) not only predict the mixing ratio but also explicitly calculate particle number concentration, thereby better representing particle size distributions. Another type of microphysical schemes is bin schemes (e.g., Clark, 1973; Geresdi, 1998; Hall, 1980; Hashino & Tripoli, 2007; Khain et al., 2015; Ogura & Takahashi, 1973; Soong, 1974), which divides hydrometeors into discrete size bins and calculates the number concentration of particles within each bin. This method allows for a more detailed representation of microphysical processes such as condensation, evaporation, and collision-coalescence by treating them separately for each size category.

Despite progress in developing microphysics parameterizations, several key challenges remain when conducting a systematic analysis of microphysics parameterizations. First, existing studies on microphysics parameterizations are scattered across a vast body of literature, making it difficult to identify mainstream schemes and development trends from individual case studies. Second, the choice of microphysics parameterizations may vary across different regions and countries, but such variations have not been quantified. Finally, microphysical parameterizations may contribute substantially to errors in forecasts and simulations of weather events, particularly in inaccurate predictions of precipitation. Although studies have been conducted to evaluate the performance of specific microphysics parameterizations, they often focus on specific weather conditions, time periods, and regions, lacking systematic comparisons across a more general breadth of locations and meteorological conditions. As a result, it remains unclear whether systematic biases exist in precipitation forecasts and simulations associated with different microphysics parameterizations.

Analyzing a large number of scientific publications systematically is a potential approach to begin to address the above challenges for three reasons. First, internet search engines for scientific literature provide a suitable means to construct large databases through systematic means to ensure a complete database for the given search criteria. Second, a large database implies a more representative sample, enhancing the generality of the results. Third, a complete literature database constructed using a systematic method allows future researchers to update that database using the exact same methods as in the original study. To our knowledge, no systematic literature review, meta-analysis, or other study has constructed a large database of literature on microphysical parameterizations, making now the perfect opportunity for an important advance in understanding such parameterizations across a large number of publications. Thus, we propose the construction of a large database of publications on precipitation forecasts and simulations.

Opportunities for advances would be ripe were such a database to exist, but manually reading thousands of publications to understand model performance and summarize systematic biases for different microphysics parameterizations would be time consuming and labor intensive. Thus, how could such a large database of studies be feasibly evaluated? The solution may lie within the newly emerging field of large language models (LLMs) (e.g., Achiam et al., 2023; Thoppilan et al., 2022; Zhao et al., 2023). LLMs process text data through attempted reading comprehension and topic classification, enabling automated analysis and understanding of literature, potentially reducing the huge manual effort in reading, classifying, and interpreting a large number of publications. Indeed, LLMs have been applied for just this purpose. For example, Callaghan et al. (2021) used an LLM to identify and classify roughly 100,000 climate-impact studies. They found that over 80% of land area over the globe had attributable anthropogenic impacts on climate and that high-income countries were twice as likely as low-income countries to have such evidence. In a second example, Miao et al. (2024) used an LLM to extract information from 314,333 hydrology publications. They were able to demonstrate trends in hydrology publications 1980–2023, showing a bias towards economically developed and densely populated regions, but leaving many river basins with frequent heavy rainfall poorly studied. These two studies highlight the potential of LLMs for extracting relevant information from a large number of publications.

The present article aims to provide an LLM-based review of the microphysical parameterizations in precipitation simulations, with the goal of offering a systematic analysis and summary for research in this field. We ask the following questions:

- What are the most commonly-used microphysics parameterizations?

- How has the usage of microphysical parameterization schemes evolved over time?

- How are different microphysical parameterizations used in different parts of the world?

- How do different microphysical parameterizations perform in simulating precipitation?

- Are there any regional variabilities in systematic biases for different microphysics parameterizations?

To address these questions, we propose the following study. First, scientific-literature search engines will be used to build a large database of scientific publications on precipitation forecasting and simulation using a mesoscale model. Second, an LLM will be used to extract information about the mesoscale model and its formulation from each publication, including the cloud microphysics parameterization and its performance. Section 2 of this article describes these two steps: the processes of building the literature database, using the LLM to extract information, and evaluating its performance through comparison to a manual extraction method. Third, we describe the characteristics of the database, including counts, authors, citations of publications, and countries and research institutions of the authors (Section 3). Fourth, we discuss the temporal distributions of usages of different microphysics parameterizations, the geographical distributions of simulation domains as well as co-occurrence relationships between microphysics parameterizations and other parameterizations (Section 4). Fifth, we show the performance of different microphysics parameterizations on precipitation simulations (Section 5). Finally, Section 6 concludes this article.

## 2 Data and Methods

This present article uses the LLM GPT-4 Turbo to extract information related to microphysical parameterizations from a large number of research publications of precipitation forecasting and simulation, which we then subsequently analyze quantitatively. The workflow is summarized in Fig. 1, and this section is organized as follows. Section 2a describes how relevant publications were collected from scientific-literature search engines to build the database (Fig. 1a). Section 2b introduces GPT-4 Turbo. Section 2c shows how to identify the relevance of each publication using GPT-4 Turbo (Fig. 1b). Section 2d describes how to extract information (such as model settings and performance) from the relevant publications through GPT-4 Turbo (Fig. 1c). Section 2e evaluates the performance of GPT-4 Turbo.

### 2.1 Literature collection

This subsection describes how to identify and select a database of journal publications (Fig. 2). First, we constructed an original literature database related to the simulation of precipitation by the Weather Research and Forecasting (WRF) model (Powers et al., 2017; Skamarock & Klemp, 2008; Skamarock et al., 2008; Skamarock et al., 2021; Skamarock et al., 2019). We chose the WRF model because it is one of a few mesoscale meteorological models to give users different options for selecting the microphysics parameterization and, as a community model with a large user base, has resulted in a large number of publications. Second, we performed specified queries on two well-known scientific-literature search engines: Web of Science and Scopus. When we did a search, we set the field tag as topic (TC) on Web of Science, and a combination of title, abstract, and key (TITLE-ABS-KEY) on Scopus (Fig. 2). The query for the WRF model was 'WRF', and the queries for the precipitation were 'precipitation', 'rain', and 'snow'. The search was conducted on 22 June 2023, containing all publications in the two search engines. After the initial search, we constructed two Microsoft Excel documents, including titles, authors, published dates, abstracts, DOIs, languages, and publication types of publications from the search results from both Web of Science and Scopus. Duplicates contained in both the Web of Science and Scopus data were removed by the DOIs. We deleted these duplicates from the Scopus database, resulting in 3883 publications from Web of Science and 803 from Scopus. We then manually excluded book chapters, conference abstracts, literature reviews, and journal publications in languages other than English. Applying these criteria, the literature database consisted of 3958 publications (3634 from Web of Science and 324 from Scopus).

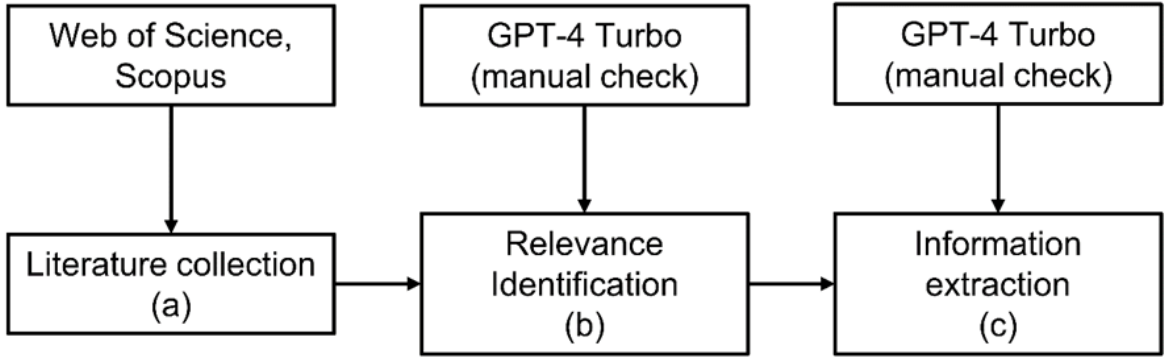

Figure 1. The framework of data collection and processing.

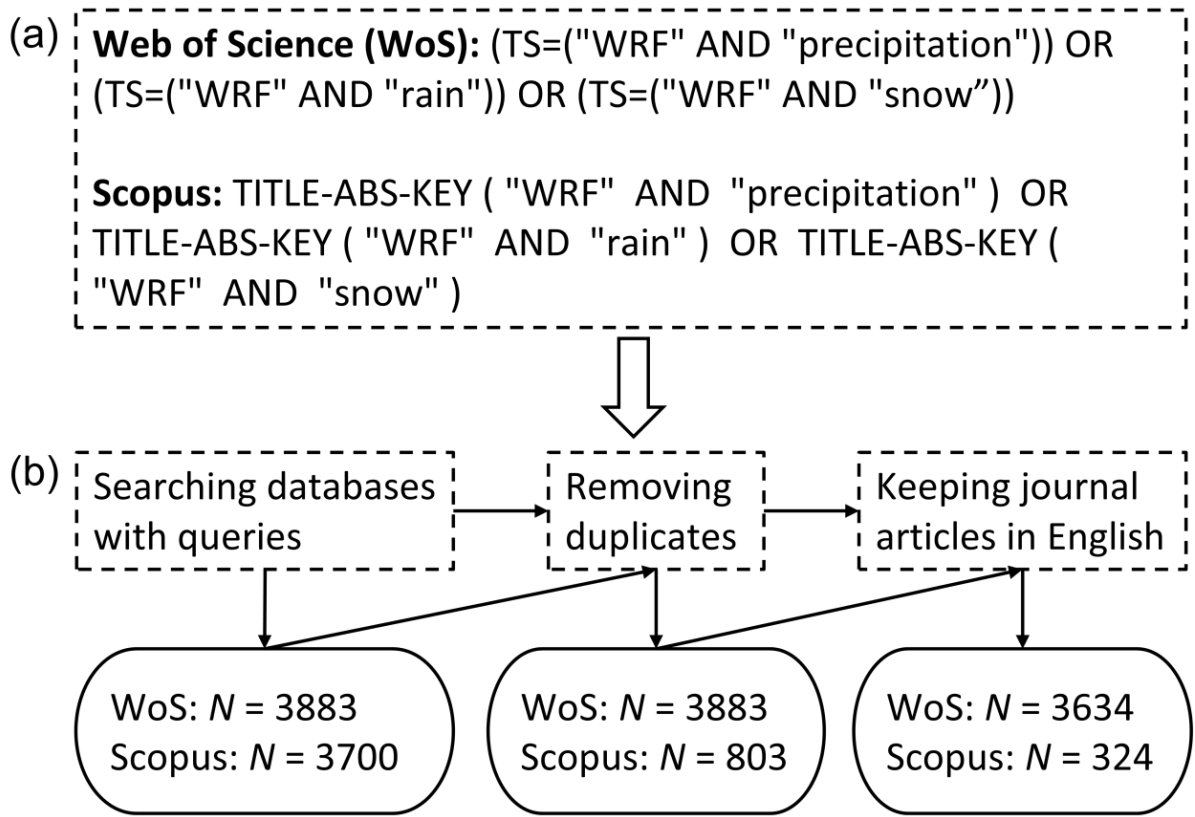

Figure 2. The process of data collection. (a) The scientific-literature search engines and keywords used to establish the literature database. (b) The three steps used to identify relevant publications, where *N* is the number of abstracts remaining at each step.

## 2.2 GPT-4 Turbo Model Overview

OpenAI's GPT-4 Turbo is an advanced LLM that belongs to the family of generative pretrained transformers (GPTs). Some recent studies showed that GPT models performed better than other traditional natural language processing models (e.g., BERT model) in text classification, inference, and question-answering tasks (e.g., Pawar & Makwana, 2022; Zhong et al., 2023). Therefore, in the present article, we employed GPT-4 Turbo due to its high intelligence and widespread application to assess the relevance of literature and to extract information. GPT models understand and generate human-like text based on the context provided. The models have been trained on a vast corpus of diverse textual data, from books, articles, and websites, across a variety of cultures, geographical locations, and human authors, which enables it to grasp a wide range of topics and language patterns. The training process of GPT models involves two main phases: pretraining and fine-tuning. During the pretraining phase, the model learns to predict the next word in a sentence by being exposed to large-scale databases (Radford et al., 2018). This phase helps the model acquire general language understanding. In the fine-tuning phase, the model is further trained on more specific databases, which can be tailored to particular domains or tasks, enhancing its performance in those areas. GPT-4 is the current fourth generation in this series of models, and GPT-4 Turbo is an advanced version, published in April 2024. Although the model parameters and optimization strategies have been continuously updated, the basic training principles and model structure have remained unchanged.

LLMs are complex systems with a multitude of parameters governing learning and text generation. Model size refers to the total number of these parameters, indicating the model's complexity and data processing capability. Although the number of parameters in GPT-4 Turbo is not officially available, Koubaa (2023) mentioned that the number of GPT-4 parameters is over 170 trillion. Therefore, the large size of GPT-4 Turbo allows it to learn complex relationships between words and phrases in training data. Meanwhile, the computational demands of such large models can reduce efficiency, prompting improvements in its optimization strategies to enhance computational efficiency during training and deployment. Based on these improvements, GPT-4 Turbo has showcased its ability to understand context and extract relevant information from vast amounts of data (Foppiano et al., 2024; Rampal et al., 2024).

When applying a GPT model to address a particular task, researchers provide background information and specific instructions or questions to guide the model. The instructions are called prompts. In our study, the collected scientific publications serve as the context, giving the model the necessary background information. We used PDFMiner, a Python library designed for extracting information from PDF documents, to parse these literature PDFs and access their contents programmatically. Then, the read information was assigned to a variable, which serves as context. Subsequently, we created prompts to direct the GPT model's attention to specific tasks including relevance identification and information extraction. The context, along with a prompt, served as the input to the GPT model. The input text was initially encoded into computer-readable tokens, forming ordered sequences. The model understands the text content through relationships between tokens and generates the answers we need.

2.3 Relevance identification

The flow of relevance identification is illustrated in Fig. 3. First, we filtered out abstracts involving WRF-Chem, WRF-CMAQ, and WRF-Hydro models through keyword searches in Microsoft Excel documents because we were only interested in precipitation simulated using WRF, not coupled models. Thus, the 3958 publications were reduced to 3559 publications. Then, we set a prompt (Fig. 4a) in GPT-4 Turbo to identify the relevance of the remaining 3559 abstracts. These abstracts were input as context into GPT-4 Turbo. If the answer to the prompt was 'yes', the publication was regarded as a relevant publication, whereas if the answer was 'no', this publication was not relevant. After screening, the literature database contained 2803 relevant publications. To validate the accuracy of GPT-4 Turbo, we chose the first 897 abstracts on the list alphabetically by document names from the literature database and identified the relevance of these abstracts manually. The manual result was 712 out of 897 (79.4%) abstracts as relevant, whereas the algorithmic result was 695 (77.5%) abstracts as relevant. If the manual and algorithmic answers were the same for a given abstract, GPT-4 Turbo was considered accurate. In total, 810 out of 897 (90.3%) GPT-4 Turbo answers were accurate.

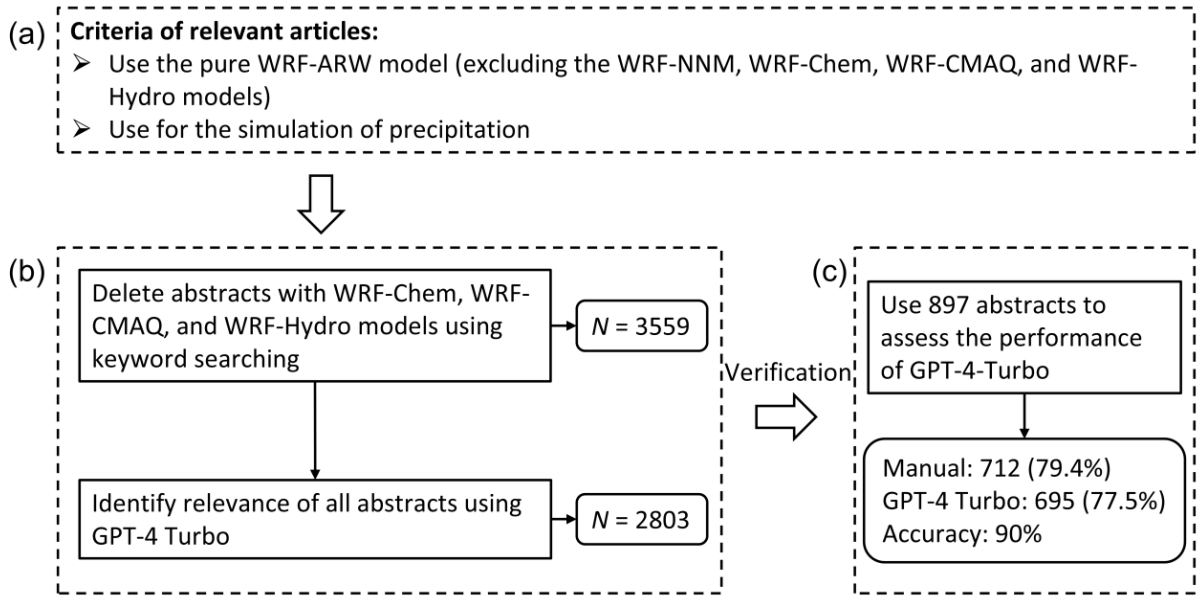

Figure 3. The process of relevance identification. (a) The criteria used to identify the relevance of publications. (b) The two steps used to identify relevant publications, where *N* is the number of abstracts remaining at each step. (c) The process of verification of GPT-4 Turbo.

2.4 Information extraction through GPT-4 Turbo

After the relevance identification, we downloaded full texts of 2803 relevant publications as PDF files. However, 104 of them were not found. Therefore, 2699 publications are the final database for the information extraction (Fig. 1c).

To extract the desired information from the publication, two steps are required. The first step is to design eight questions related to microphysics parameterizations and precipitation simulations. To better understand the systematic biases of microphysics parameterizations, we need to know the configurations of the WRF model, simulation domain, and performance of microphysics parameterizations from each publication. The questions we asked in the prompt are shown in Fig. 4b. To ensure GPT-4 Turbo best captures results from different synonyms, we listed as many descriptors as we could. For example, Question 1 is about configurations of WRF model, so we detailed the type names of the five main parameterizations that are common in the configurations. Questions 4, 5, and 6 are about performance of the simulations. In the three questions, 'overestimation', 'underestimation', and 'precipitation' can be expressed in many ways in publications. We acknowledge that models do not estimate precipitation—they simulate or predict it—but, most articles tend to use "estimate" rather than other more precise terms. Thus, we stick with this terminology throughout the present article. The second step is to supply the 2699 publications to GPT-4 Turbo as context and then to ask GPT-4 Turbo these eight questions (Fig. 4b).

When applying GPT-4 Turbo, there are a number of different parameters that can be set by the user. We left those at their default values, except for one: the temperature parameter to control the randomness of text generation. The temperature parameter ranges from 0 to 2, where a lower temperature parameter (closer to 0) results in more deterministic outputs, with the model tending to choose the most likely word or token. In contrast, higher values such as 0.8 will make the output more random. We chose a setting of 0.12 after several tests of different temperature parameters because it reached a balance between generating deterministic and creative outputs and resulted in relatively accurate answers.

The output of GPT-4 Turbo was written in ASCII text format and then transferred into Microsoft Excel documents. This approach was convenient for the subsequent quantitative analysis.

**a. Input prompt for relevance identification**

Does this study apply the WRF model for precipitation simulation (Here, Precipitation includes rainfall and snowfall)? Additionally, if the study is about precipitation simulation and does not explicitly mention the model used in the abstract, consider it as an application of the WRF model. Please analyze the text and provide answers to the above questions. Please just answer Yes or No.

**b. Input prompt for information extraction**

Please answer the following questions based on the study provided. Note: Each question should be considered independently, and the documentation should be reviewed thoroughly as if each question was asked one by one.

1. What was the configuration/setup of physical parameterizations for each WRF application? (The set of physical parameterization options including cloud microphysics, radiation, cumulus, boundary layer schemes)? And what was the land surface model used? (Please list the names directly)

2. What is the simulation domain or study area for each WRF application? Please answer this question directly without additional context or explanations.

3. For each WRF application, which precipitation-related variable was evaluated: accumulated precipitation/rainfall amount, instantaneous precipitation/rainfall rate, or reflectivity? If it was accumulated precipitation amount, specify the accumulation period. Please answer this question directly without additional context or explanations.

4. For each WRF application, was the simulated precipitation overestimated or underestimated compared to observed data? Please describe the corresponding WRF configuration. The 'overestimated' can be replaced by wet bias, positive bias, more, more intense, heavier, larger, higher, wetter, stronger, exceeds, produce an excess of rainfall, or produce too much precipitation, etc. The 'underestimated' can be replaced by dry bias, negative bias, less, less intense, weaker, smaller, lower, drier, generate too little precipitation, fail to simulate or cannot simulate precipitation, etc. Provide a summary.

5. Where the precipitation was overestimated compared with observed data? Note please list regions. (The 'overestimated' can be replaced by wet bias, positive bias, more, more intense, heavier, larger, higher, wetter, stronger, exceeds, produce an excess of rainfall, or produce too much precipitation, etc. The 'precipitation' can be replaced by rainfall, rain amount, or P, etc.). Provide a summary.

6. Where the precipitation was underestimated compared with observed data? Note please list regions. (The 'underestimated' can be replaced by dry bias, negative bias, less, less intense, weaker, smaller, lower, drier, generate too little precipitation, fail to simulate or cannot simulate precipitation, etc. The 'precipitation' can be replaced by rainfall, rain amount, or P, etc.). Provide a summary.

7. Are there any RMSE (Root Mean Squared Error) values for each WRF application? If so, list the values and units one by one. Please review all text and tables thoroughly. Note 'yes' or 'no' first, then provide all values from all text and tables. If there are different WRF configurations, list the error values along with their corresponding microphysics scheme. Please ensure that your answers are accurate and complete.

8. Are there any correlation coefficients (CC) for each WRF application? If so, list the values one by one. Please review all the text and tables.

Figure 4. The prompt used in the present article. (a) Input prompt questions for relevance identification. (b) Input prompt questions for information extraction, numbered 1 through 8.

2.5 Evaluating the performance of GPT-4 Turbo

To validate the information extracted, we selected 300 publications to manually check the accuracy of the answers produced by GPT-4 Turbo. The lead author read each of the 300 publications independently and wrote down the answers to the eight questions. The manual answers were double-checked and then were compared by the lead author with GPT-4 Turbo's answers. In every situation, whether GPT-4 Turbo got the right answer or not was unambiguous. Table 1 shows the effectiveness of extracting WRF model configuration and performance information from each publication using GPT-4 Turbo, demonstrating excellent performance with an average accuracy of 94% for all eight questions. Specifically, question 1 on model configuration had the highest accuracy of 98%, but question 4 on the performance of the WRF model had the lowest accuracy of 86%.

Through comparing manual and GPT-4 Turbo answers, wrong answers were divided into two types. One was that GPT-4 Turbo provided irrelevant answers or did not provide answers at all. These wrong answers may be caused by systematic biases of GPT-4 Turbo because LLMs have hallucination problems (i.e., the frequency at which a language model generates incorrect or nonsensical information). The other was that GPT-4 Turbo provided incomplete answers, particularly for questions 7 and 8 associated with root mean squared errors (RMSEs) and correlation coefficients (CCs). In these situations, some publications studied more than one case using different configurations of WRF, or in different regions, or during different study periods. Therefore, more than one or two RMSE or CC values appeared. Also, GPT-4 Turbo had difficulty finding these two metrics in the publications if the metrics appeared, not within the text, but within tables or figures. Given these challenges, the high accuracies for GPT-4 Turbo provided fairly reliable information for our study overall.

Table 1. The accuracy of the eight questions in Fig. 4 from a comparison of manual answers versus answers from GPT-4 Turbo.

| Question number | Name of question | Accuracy of GPT-4 Turbo |
|---|---|---|
| 1 | Microphysics parameterization | 96% |
| | Radiative parameterization | 98% |
| | Cumulus parameterization | 98% |
| | Planetary boundary layer parameterization | 98% |
| | Surface layer parameterization | 98% |
| 2 | Location of simulated domain | 95% |
| 3 | Variable associated with precipitation | 94% |
| 4 | Performance | 86% |
| 5 | Overestimated region | 94% |
| 6 | Underestimated region | 93% |
| 7 | RMSE | 92% |
| 8 | CC | 90% |

**3 Results of the machine-assisted literature review**

In this section, we examine the characteristics of the 2803 relevant publications, focusing on the analysis of publication counts, authorship, and citations in Section 3a. Additionally, Section 3b delves into an analysis of the countries and affiliations associated with the publications.

3.1 Counts, authors, and citations of publications

The blue line in Fig. 5a shows the number of publications associated with precipitation simulations by year of publication 2003 to 2023. The annual number of relevant publications increased from 1 to more than 300 with time. The number of relevant publications in 2023 was much less than that in 2022 because the end of the study period was June 2023, and the data in 2023 was not complete. The rate of increase in the annual number of relevant publications was slower before 2007 (with an average growth rate of about 3 publications per year) than after 2007 (with an average growth rate of about 20 publications per year). This trend reflects the increasing use and popularity of the WRF model for precipitation forecasting and simulation. There are three reasons for its popularity. First, the WRF model is open source, which allows researchers and institutions to freely access, use, and modify the models. Second, the WRF model has a large user community and rich documentation resources. In addition, regularly updated and improved model versions also provide the latest techniques and methods. Third, the WRF model has many configuration options (including initial and boundary input data, resolutions, various parameterizations). Users can select and combine different configurations to improve the accuracy of precipitation simulations. The orange line in Fig. 5a shows the temporal distribution of the average number of authors per publication. Prior to 2008, the average number of authors per publication fluctuated, ranging from two to five, which can be attributed to the relatively low number of publications on WRF model simulations of precipitation in earlier years. However, after 2008, the average number of authors per publication stabilized between four and five, showing an upward trend. This trend reflects an increasing level of collaboration among researchers.

We examined the number of relevant publications per journal, and the mean and median Web of Science citation count per journal (Fig. 5b). This analysis not only helps us identify the leading journals in the field of precipitation simulation but also highlights those that have garnered wider attention and higher citation counts. Figure 5b presents the top 15 journals with the highest number of publications related to precipitation simulations. Specifically, *Journal of Geophysical Research-Atmospheres* and *Monthly Weather Review* lead with the most publications, boasting 192 and 190 publications, respectively, followed by *Atmospheric Research* with 183 publications. Notably, six of these top 15 journals are affiliated with the American Meteorological Society (AMS). Specifically, *Monthly Weather Review*, *Journal of Climate* and *Weather and Forecasting* rank the top three in both the mean and median citation counts, emerging as the most influential journals. *Monthly Weather Review* has the highest mean of 58 citations per publication but the smaller median of 19 citations per publication because this journal has two publications with over one thousand citations (Morrison et al., 2009; Thompson et al., 2008). This is followed by *Journal of Climate* (with a mean of 51 and median of 24 citations per publication) and *Weather and Forecasting* (with a mean of 40 and median of 21 citations per publication).

3.2 Countries and research institutions of authors

This subsection analyzes countries of authors and research institutions of 2803 publications. We excluded 205 publications that did not record research institutions and countries of authors, leaving 2598 publications for analysis in this subsection. Figure 6a shows the number of publications on precipitation simulations by different countries. The United States and China dominate the number of publications, with 1067 and 795 publications, respectively, accounting for the majority of the research output. In addition to the United States and China, India, Germany, and South Korea have also published on precipitation simulations using WRF (252, 120, and 105 relevant publications, respectively).

Figure 6b shows the top 20 research institutions from different countries based on the number of publications related to precipitation simulations. Among these institutions, the Chinese Academy of Sciences ranks first with 545 publications, more than twice as many as the number of publications by National Center for Atmospheric Research (NCAR), which has 254 publications. Notably, of the top 20 institutions, eight are from the United States and seven are from China. However, despite the United States having more institutions in the top 20, China has published 360 more articles than the United States in this field. This difference in publication numbers highlights key characteristics regarding the distribution of research resources in both countries. China's overall advantage in publication volume can largely be attributed to the concentrated efforts of the Chinese Academy of Sciences in rainfall simulation research. In contrast, the United States shows a more decentralized distribution of research resources, indicating that the U.S. has a broad network of research institutions contributing to rainfall simulation, including NCAR, NASA, NOAA, and various universities.

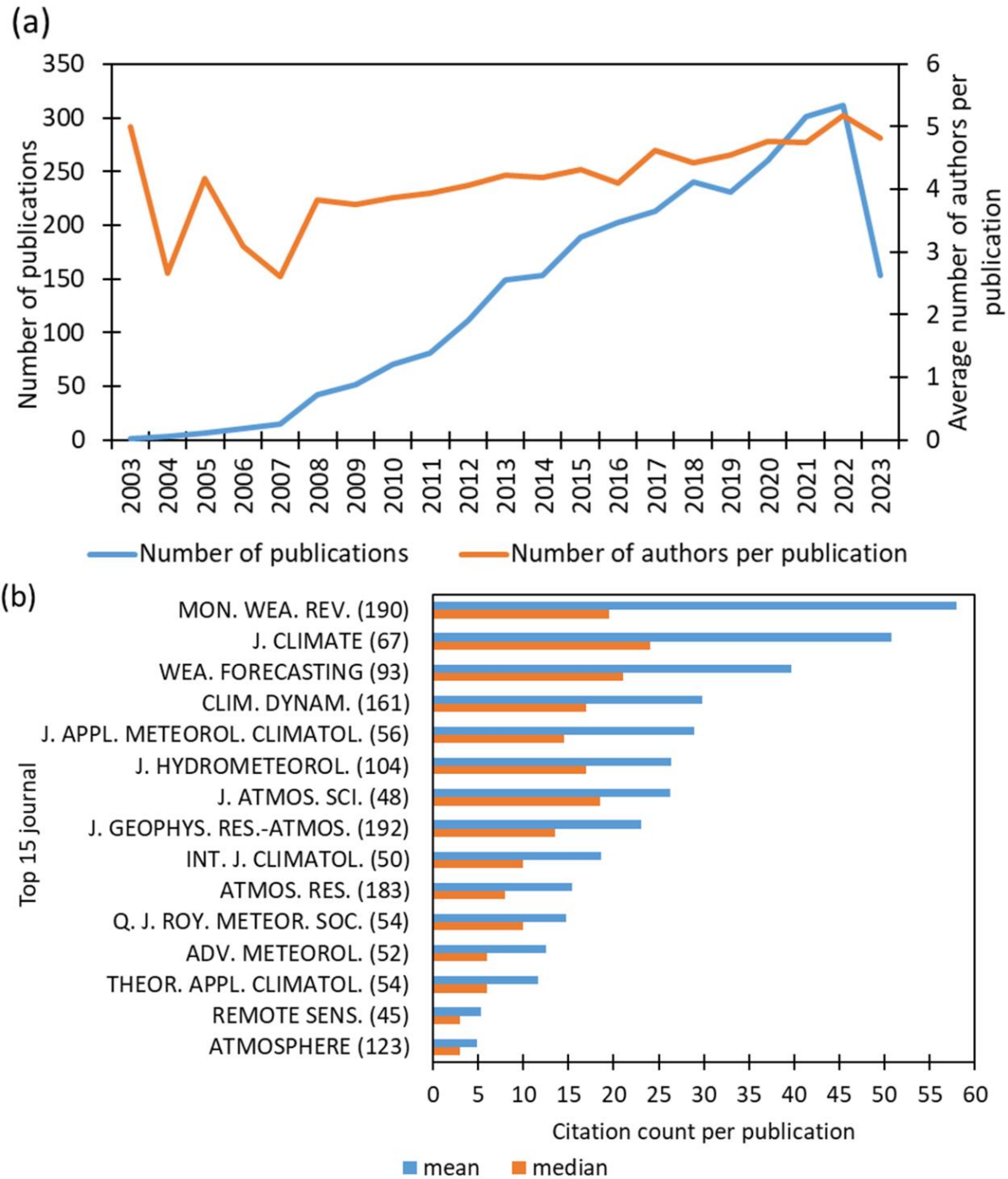

Figure 5. (a) The number of publications associated with precipitation simulations and the average number of authors per publication from 2003 to 2023, (b) the mean and median Web of Science citation counts per publication in top 15 journals. Numbers in parentheses are the total number of publications per journal.

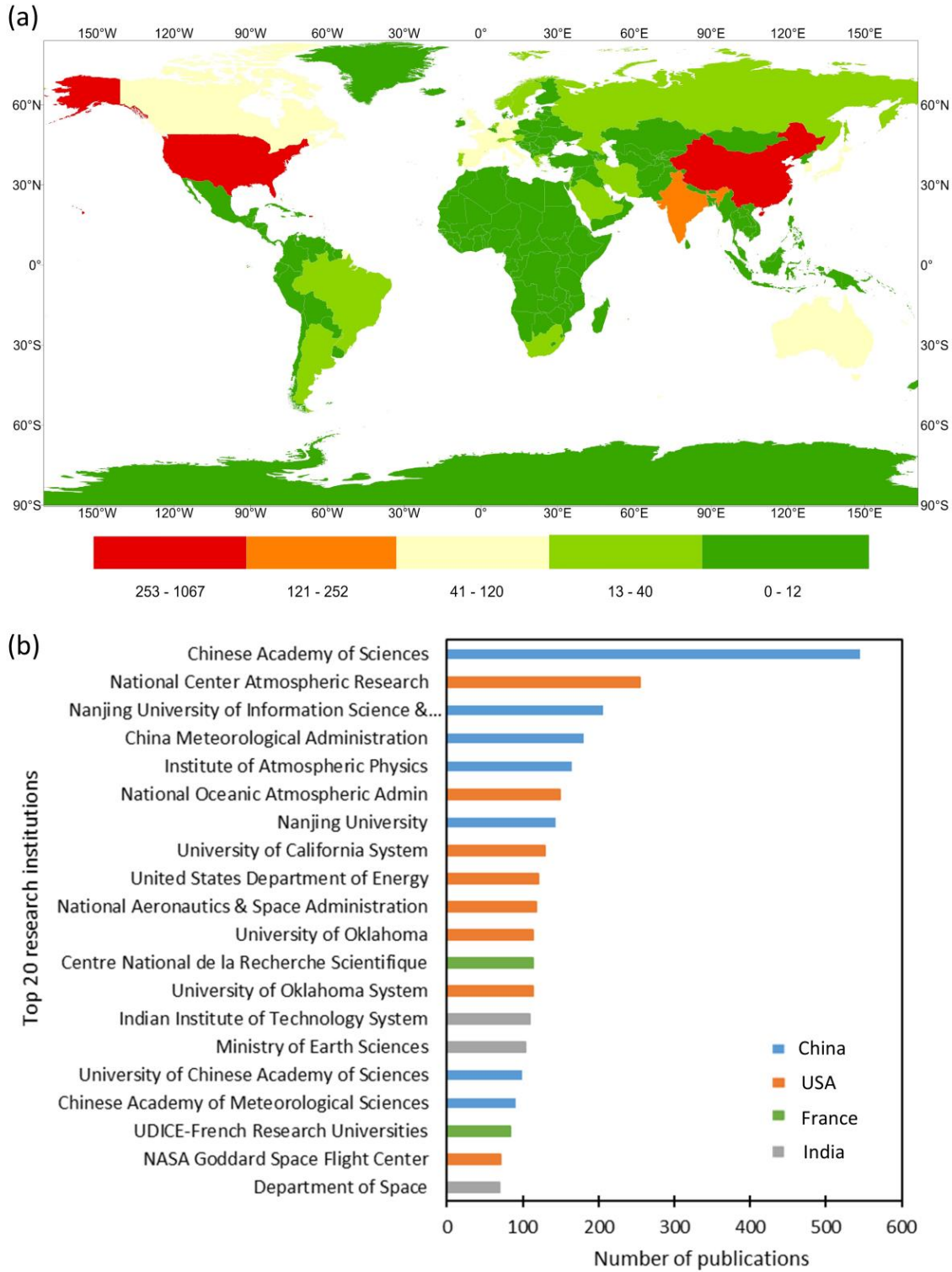

Figure 6. The number of publications (a) in different countries, (b) in the top 20 research institutions.

**4 Characteristics of different bulk microphysics parameterizations**

In this section, we study the characteristics of microphysics parameterizations, including the temporal distribution of the simulations (section 5a), the geographical distribution of the simulation domains (section 5b) using different microphysics parameterizations, and the co-occurrence between microphysics parameterizations and other types of parameterizations (section 5c).

4.1 Temporal distribution of simulations using different bulk microphysics parameterizations

Microphysics parameterizations are divided into bulk parameterizations (characterizing the particle size distribution through semi-empirical functions) and bin parameterizations (allowing for the free evolution of the particle size distribution). Bulk microphysics parameterizations include the WRF Single-Moment 6-class (WSM6; Hong & Lim, 2006), Thompson (Thompson & Eidhammer, 2014; Thompson et al., 2008), Morrison (Morrison et al., 2009), Lin (Chen & Sun, 2002), WRF Single-Moment 5-class (WSM5; Hong et al., 2004), WRF Double-Moment (WDM: WDM5, WDM6, WDM7; Lim and Hong, 2010), WRF Single-Moment 3-class (WSM3; Hong et al., 2004), Goddard Cumulus Ensemble (GCE; Tao & Simpson, 1993), Ferrier (Rogers et al., 2001), Milbrandt–Yau (MY; Milbrandt & Yau, 2005), Kessler (Kessler, 1969), National Severe Storms Laboratory (NSSL; Mansell et al., 2010), Stony Brook University Lin–Colle (SBU-Lin; Lin & Colle, 2009), Predicted Particle Properties (P3; Morrison & Milbrandt, 2015), and Community Atmosphere Model Morrison–Gettelman (CAM; Morrison & Gettelman, 2008) parameterizations. Due to a small number of publications using each category of bin parameterization, we grouped them into one category, called bin parameterizations. Microphysics parameterizations used in fewer than ten publications during the study period were not considered further. Whereas 95% of publications used bulk microphysics parameterizations to simulate precipitation, only 5% used bin parameterizations (Fig. 7a). Because bulk parameterizations tend to be cheaper and more efficient for computation, bulk parameterizations tend to be more popular than bin parameterizations (e.g., Hu & Igel, 2023; Khain et al., 2015; Morrison et al., 2020). Therefore, we focus on bulk microphysics parameterizations in the rest of this article. The most popular bulk microphysics parameterization was the WSM6 parameterization (used in 776 publications), followed by the Thompson parameterization (used in 678 publications) (Fig. 7a). These two parameterizations are used in publications more than twice as frequently as other microphysics parameterizations.

Figure 7b shows the distribution of different microphysics bulk parameterizations by year from 2003 to 2023. The number of publications using one-moment parameterizations increased with time until about 2017 when it stayed steady at about 160 publications per year. Before 2020, one-moment parameterizations were dominant in precipitation simulations, comprising 100% in 2004 to 51% in 2020. Among these one-moment parameterizations, the Lin parameterization was the most popular before 2007, but after this year, the WSM6 parameterization became the most popular one. To explain this change, the source code for the early one-moment bulk parameterization (i.e., Lin) was added into the WRF code in 2000, but publications using the parameterization started to appear after 2004. The WSM6 parameterization's source code was added into WRF code in 2004 and first used in publications associated with precipitation simulation two years later.

The number of publications using double-moment parameterizations increased through the study period. Double-moment parameterizations' sources codes were officially added into WRF code beginning in 2008 and they were used in precipitation simulations beginning in 2009. In 2021, the double-moment parameterizations became the dominant type (53%). Among these double-moment parameterizations, the Thompson parameterization was the most popular parameterization from 2020, first added to the WRF code in 2009. The NSSL parameterization is a collection of one-moment and double-moment versions. This parameterization was added to WRF in 2012 but used in the precipitation simulations beginning in 2014. This parameterization was not popular in precipitation simulations, used in fewer than 10 publications per year. Overall, publications using a new parameterization usually began appearing one or two years after the parameterization source code was added to the WRF code. The lag is more likely due to the time needed to conduct the research using a new scheme and then write a manuscript about it (and have it published).

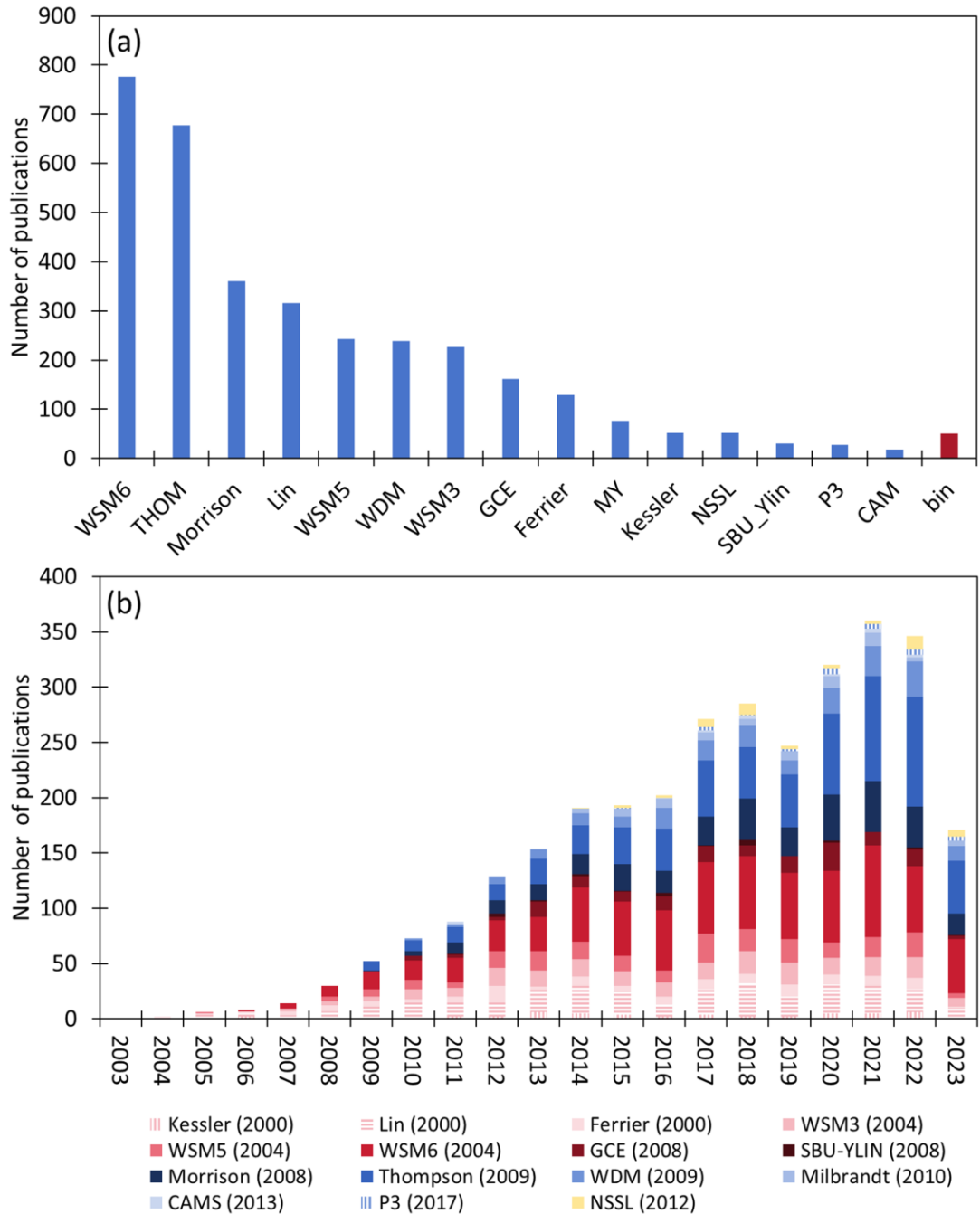

Figure 7. (a) The distribution of the number of publications for different categories of microphysical parameterizations (blue bars for bulk microphysics parameterizations, red bar for bin microphysics parameterizations). (b) The time series of the number of publications using different bulk microphysical parameterizations (the number in parentheses is the year when the parameterization's code was officially added into the WRF code). Red bars represent single-moment bulk microphysics parameterizations, whereas blue bars represent double-moment bulk microphysics parameterizations.

4.2 Geographical distribution of simulation domains using different microphysics parameterizations

To understand the geographical distribution of simulation domains in publications associated with precipitation simulations, question 2 asked about the simulation domain in the prompt. The answer from GPT-4 Turbo gave the name of region (e.g., southeast China, central United States, or Oklahoma) for the simulation domain in each publication. Then, we asked GPT-4 Turbo the northern and southern latitudes, and western and eastern longitudes of this domain, which went into Fig. 8 showing the geographical distribution of the number of publications related to the WRF-simulated precipitation.

China (particularly southeast China and the Sichuan Basin) and the central United States have the most publications (more than 220 in each region in total). There are several potential reasons. First, extreme rainfall events coupled with considerable social and economic impacts often occurred in these regions, such as typhoons over western North Pacific (Weinkle et al., 2012) and tornadoes in the central United States (Goliger & Milford, 1998; Maas et al., 2024). Second, there are large WRF user communities in China and the United States. The Himalayas, Korean Peninsula, and their surroundings also had many publications (more than 200 in each region).

Figure 9 shows the geographical distribution of the number of publications using the nine main microphysics parameterizations. These microphysics parameterizations have a similar geographical distribution to that in Fig. 8, with notable popularity in eastern Asia, southern Asia, and the United States (Fig. 9). However, differences among microphysical parameterizations exist. The Lin parameterization was predominantly used to simulate precipitation in eastern India (over 30), the Sichuan Basin (over 30), and Zhejiang Province (over 36; Fig. 9a). Both the Ferrier and GCE parameterizations shared a similar geographical distribution, with the highest number (over 18) of publications in the central United States (Figs. 9b,c). The regions where the most publications used the WSM3 and WSM5 parameterizations were the Sichuan Basin (over 30) and southeastern China (over 36 near Taiwan), respectively (Figs. 9d,e). The WSM6 parameterization was widely adopted in the central United States and southern China (over 78; Fig. 9f). Compared with other parameterizations, the WSM6 parameterization was the most popular over the globe (Figs. 9f). For the WDM parameterization, the Korean Peninsula was the most popular region followed by the Tibetan Plateau and south Asia (Fig. 9g). The Thompson and Morrison parameterizations were more popular in the United States, particularly the southern part of the central United States with the maximum number of publications of 60 and 42, respectively (Figs. 9h,i). Interestingly, precipitation in Antarctica was simulated using the Morrison and WDM parameterizations.

Overall, the Lin, WSM3, WSM5, and WDM parameterizations were more frequently used in southern Asia and China, whereas the Ferrier, Thompson, and Morrison parameterizations were more frequently used in the United States. The numbers of publications using the WSM6 parameterization were comparable between the United States and China.

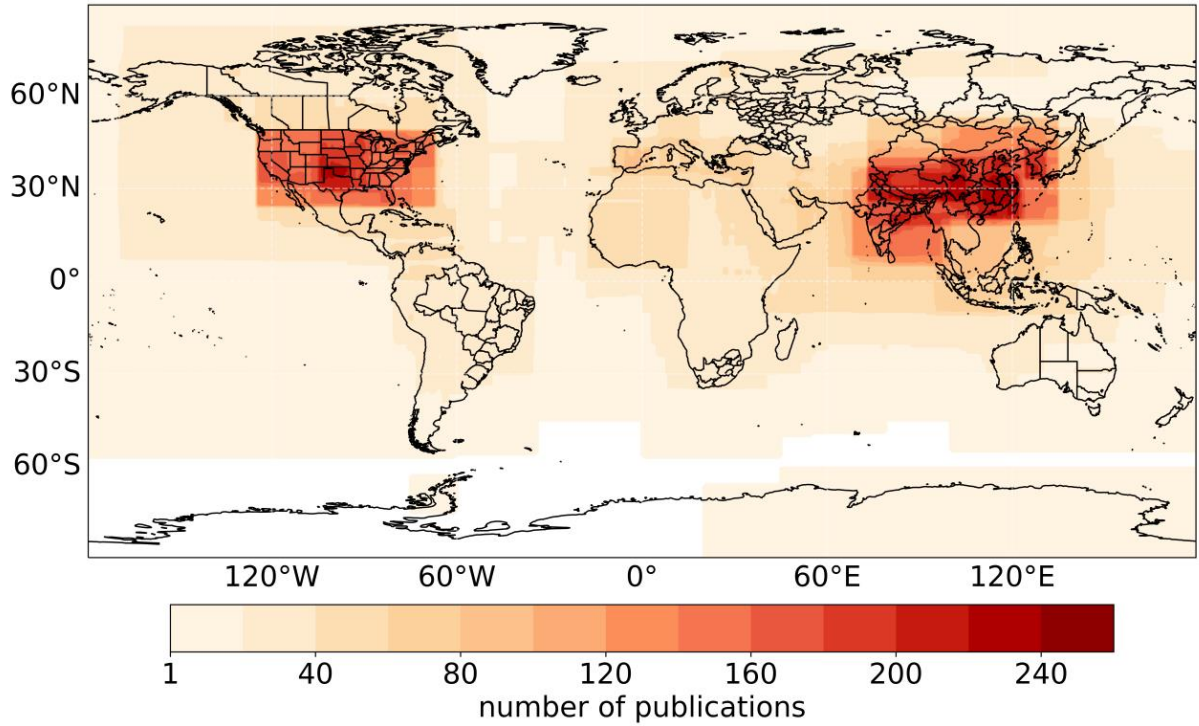

Figure 8. Global map of the number of publications associated with WRF-simulated precipitation (the location is the simulation domain in each publication).

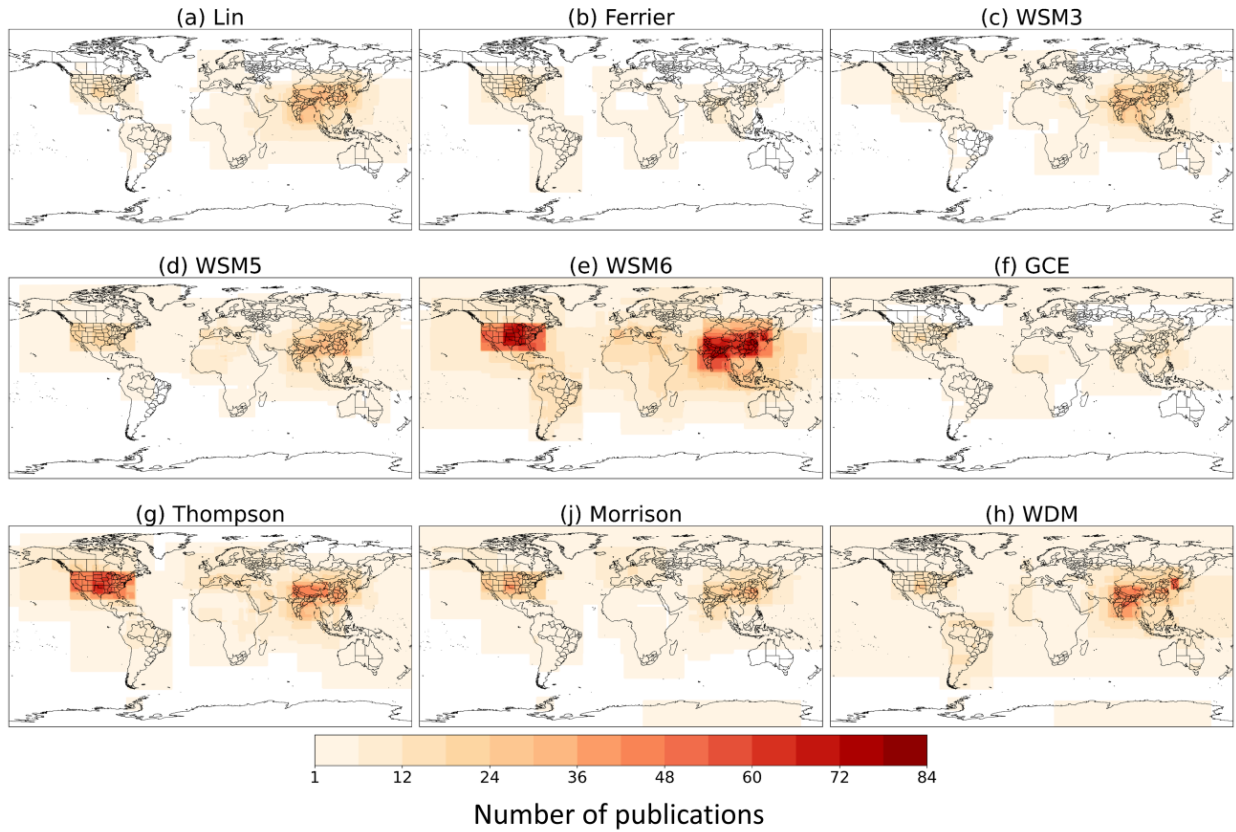

Figure 9. Global map of the number of publications associated with WRF-simulated precipitation using (a) Lin, (b) Ferrier, (c) GCE, (d) WSM3, (e) WSM5, (f) WSM6, (g) Thompson, (h) Morrison, and (i) WDM parameterizations (the location is the simulation domain in each publication). The first two rows contain one-moment microphysics parameterizations, and the third row contains double-moment parameterizations. The orders of panels (a)–(f) and (g)–(h) are based on the year when the microphysics parameterization was added to the WRF model.

4.3 Co-occurrence relationships between microphysics parameterizations and other parameterizations

To find which configurations are popular when using different microphysics parameterizations to simulate precipitation, this subsection discusses the relationship between microphysics parameterizations and those for other sub-grid-scale processes (cumulus, planetary boundary layer, radiation, surface parameterizations). In the 2699 publications, 6394 sets of configurations of the five physics parameterization types listed above were used to simulate precipitation (i.e., some publications used more than one configuration in the WRF model). In these configurations, parameterizations that were used 500 times or fewer were relegated to 'other' (and not further analyzed). Table 2 shows the percentages of the two most-used land-surface parameterizations (Noah; RUC), three most-used planetary boundary parameterizations (Yonsei University, YSU; Mellor–Yamada–Janjić, MYJ; Mellor–Yamada–Nakanishi–Niino, MYNN), three most-used cumulus parameterizations (Betts–Miller–Janjić, BMJ; Kain–Fritsch, KF; G, including Grell–Devenyi, Grell–Freitas, Grell-3), and three most-used longwave/shortwave radiation parameterizations (Rapid Radiative Transfer Model/Dudhia, RRTM/Dudhia; Rapid Radiative Transfer Model for GCMs, RRTMG; NCAR Community Atmosphere Model, CAM). The most popular land-surface parameterization was Noah (65.8%) (Table 2). The most popular PBL parameterization was YSU (48.2%), followed by MYJ and MYNN (Table 2). The choice of using a cumulus parameterization or not is typically based on the horizontal grid spacing of the model simulation. A cumulus parameterization is typically used in a simulation with horizontal grid spacing more than 4 km; otherwise, it is not used. The KF and G parameterizations were the most popular (37.8% and 22.2%, respectively; Table 2). Finally, Table 2 shows that the RRTM/Dudhia parameterization was the most commonly used radiation parameterization (39.2%), followed by the RRTMG parameterization (22.4%).

Although there were 6394 configurations used in the simulations in our database, some of these configurations were quite popular. The top three most popular configurations for each microphysics parameterization appear in Table 3. Regardless of the choice of microphysics parameterizations, the configuration featuring the RRTM/Dudhia, KF, YSU, and Noah parameterizations was the most popular within our database, appearing within 500 configurations (Table 3). However, the second most popular configuration for different microphysics parameterizations differed (Table 3). For the Lin, Ferrier, and WSM3 parameterizations, the number of publications using the configuration with the RRTM/Dudhia, BMJ, YSU and Noah parameterizations were 30, 17, 32, respectively. For the WSM5 parameterization, the number of publications using the configuration with the CAM, KF, YSU, and Noah parameterizations was 35. For the GCE and Thompson parameterizations, their second most popular configurations with RRTM/Duhdia, KF, MYJ, Noah parameterizations were the same (18 and 33 publications using this configuration, respectively). In addition, the Morrison and WDM parameterizations also have the same second most popular configuration (30 and 15 publications using this configuration, respectively). To provide a guide for the choice of parameterizations, we will show the performance of the most popular configurations with different microphysics parameterizations in the next section.

Table 2. The percentage of main parameterizations to the 6397 configurations. "None" means no parameterization was used or mentioned in publications. "Other" means parameterizations other than those listed in the text were used.

| Radiation parameterization | RRTM/Dudhia | RRTMG | CAM | Other | None |
|---|---|---|---|---|---|
| | 39.2% | 22.4% | 8.2% | 19.1% | 11.1% |
| Cumulus parameterization | KF | G | BMJ | Other | None |
| | 37.8% | 22.2% | 16.4% | 9.1% | 14.5% |
| Planetary boundary parameterization | YSU | MYJ | MYNN | Other | None |
| | 48.2% | 23.3% | 8.1% | 12.8% | 7.5% |
| Land surface parameterization | Noah | RUC | Other | None | |
| | 65.8% | 7.9% | 8.9% | 17.4% | |

Table 3. The number of publications using the top three popular configurations for each microphysics parameterizations.

| Lin | RRTM/Dudhia-KF-YSU-Noah | RRTM/Dudhia-BMJ-YSU-Noah | RRTM/Dudhia-G-YSU-Noah |
|---|---|---|---|
| | 67 | 30 | 28 |
| Ferrier | RRTM/Dudhia-KF-YSU-Noah | RRTM/Duhia-BMJ-YSU-Noah | RRTM/Duhia-G-YSU-Noah |
| | 25 | 17 | 15 |
| WSM3 | RRTM/Dudhia-KF-YSU-Noah | RRTM/Duhia-BMJ-YSU-Noah | RRTM/Duhia-G-YSU-Noah |
| | 57 | 32 | 27 |
| WSM5 | RRTM/Dudhia-KF-YSU-Noah | CAM-KF-YSU-Noah | RRTM/Duhia-G-YSU-Noah |
| | 46 | 35 | 19 |
| WSM6 | RRTM/Dudhia-KF-YSU-Noah | RRTM/Duhia-G-YSU-Noah | RRTMG-KF-YSU-Noah |
| | 137 | 53 | 53 |
| GCE | RRTM/Dudhia-KF-YSU-Noah | RRTM/Duhia-KF-MYJ-Noah | RRTMG-KF-YSU-Noah |
| | 21 | 18 | 7 |
| Thompson | RRTM/Dudhia-KF-YSU-Noah | RRTM/Duhia-KF-MYJ-Noah | RRTMG-KF-YSU-Noah |
| | 69 | 33 | 31 |
| Morrison | RRTM/Dudhia-KF-YSU-Noah | RRTMG-KF-YSU-Noah | RRTM/Duhia-G-YSU-Noah |
| | 39 | 30 | 22 |
| WDM | RRTM/Dudhia-KF-YSU-Noah | RRTMG-KF-YSU-Noah | CAM-KF-YSU-Noah |
| | 39 | 15 | 14 |

## 5 Performance of different bulk microphysics parameterizations

The aim of this section is to identify any systematic biases of different microphysics parameterizations for the precipitation simulations through collecting answers to questions 4–8 (Fig. 4) in the 2699 publications. We focus on the accumulated precipitation, instead of instantaneous precipitation rate or radar reflectivity, because it appears within 65% of the publications. We divided the answers from GPT-4 Turbo into four types: no answer, overestimation, underestimation, and mix. No answer means that GPT-4 Turbo provided *Nah* in the answer. Overestimation or underestimation means GPT-4 Turbo summarized a general tendency of overestimated or underestimated precipitation for each microphysics parameterization. Mix means that GPT-4 Turbo presented both overestimated and underestimated precipitation outcomes for each microphysics parameterization in publications due to different configurations, locations, times and without providing a general evaluation. In the present article, we just consider overestimation and underestimation answers and ignore mix answers. This is because mix answers occupied a small portion (18%), and we cannot get the proportion of overestimation and underestimation from the mix answers, making it difficult to quantify.

In total, the 1325 out of 2699 publications provide overestimation or underestimation answers for nine main microphysical parameterizations (Fig. 10a). From the distribution of the number of publications, systematic biases of each microphysical parameterization could be summarized. For WDM and GCE parameterizations, the numbers of publications with underestimation answers are 5.2% and 1.4% more than those with overestimation answers. Thus, these two parameterizations slightly underestimate the precipitation. For the remaining seven parameterizations, the number of publications with overestimation answers was larger than that with underestimation answers, suggesting a tendency for overestimation in the precipitation. Specifically, the WSM3 and Lin parameterizations have the largest percentages of the numbers of publications with overestimation (71.7% and 66.1%, respectively).

For the most popular configurations, the tendency for overestimation disappeared (Fig. 10b). Except for the WDM parameterization with a tendency to underestimate precipitation (92.5%) and the Morrison and WSM5 parameterizations with tendencies to overestimate precipitation (about 60%), the numbers of publications overestimating and underestimating were comparable for the other six parameterizations (Fig. 10b). This finding illustrates nearly all (7 out of 9) microphysics parameterizations tended to overestimate, but the degree of overestimation was reduced in the most popular configurations. Thus, whether it was consciously or unconsciously, researchers tended to choose the most popular configurations because there tended to be some perceived performance benefit in doing so.

Overall, microphysics parameterizations tended to overestimate the precipitation. Some previous case studies have investigated why the parameterizations overestimated precipitation. For example, the prediction of ice-phase particles and relevant microphysical processes strongly affects the accuracy of precipitation simulations and may explain the tendency for overprediction. McMillen and Steenburgh (2015) have reported that overestimation in precipitation using the WDM6 parameterization was associated with a large production of graupel and a small production of snow. Also, Lin and Colle (2009) have shown that the Lin parameterization overestimated surface precipitation along the Cascade windward slopes due to an increase in the conversion of snow to graupel and unrealistically rapid fallout of graupel. However, whether these kinds of studies can generalize to a larger set of publications to explain the tendency for different microphysics parameterizations to overestimate precipitation is unclear.

Due to these publications studying precipitation over various regions, we hypothesize that the systematic biases of the same microphysical parameterization might change in the different regions. To test this hypothesis, Fig. 11 shows the geographical distribution of the difference   between numbers of publications reporting overestimation and underestimation ($N_{overestimation} - N_{underestimation}$), based on the domains of the model simulations from each publication, for different microphysical parameterizations. Principal findings are as follows:

·       Simulations using the Lin, Ferrier, and GCE parameterizations tended to underestimate the precipitation in most places over the globe, except for the overestimation over India and Tibet Plateau in the Lin simulation (Figs. 11a,b,f).

·       Simulations using the WSM3 and WSM5 parameterizations had a similar distribution of systematic biases to each other (Figs. 11c,d). Both had overestimation over east Asia and southeast Asia, and both had underestimation over the eastern United States (Figs. 11c,d). The difference between these two parameterizations was the underestimation near the Mediterranean Sea in the WSM5 simulations and over Norway in the WSM3 simulations (Figs. 11c,d).

·       Most of the simulations using the WSM6 parameterization overestimated precipitation over southeast China and the northeast United States (Fig. 11e).

·       Simulations using the Thompson parameterizations tended to overestimate precipitation, particularly over the Tibetan Plateau and the western United States but tended to underestimate precipitation over the eastern United States (Fig. 11g).

·       Simulations using the Morrison and WDM parameterizations (Figs. 11h,i) overestimated precipitation occurring over China, particularly the Tibetan Plateau.

Overall, except simulations using the Lin, Ferrier, and GCE parameterizations that tended to underestimate over almost all locations, the remaining six parameterizations tended to overestimate precipitation. Specifically, precipitation over China, southeast Asia, the western United States, and central Africa tended to be overestimated, whereas precipitation over the eastern United States tended to be underestimated by the WRF model using most microphysics parameterizations.

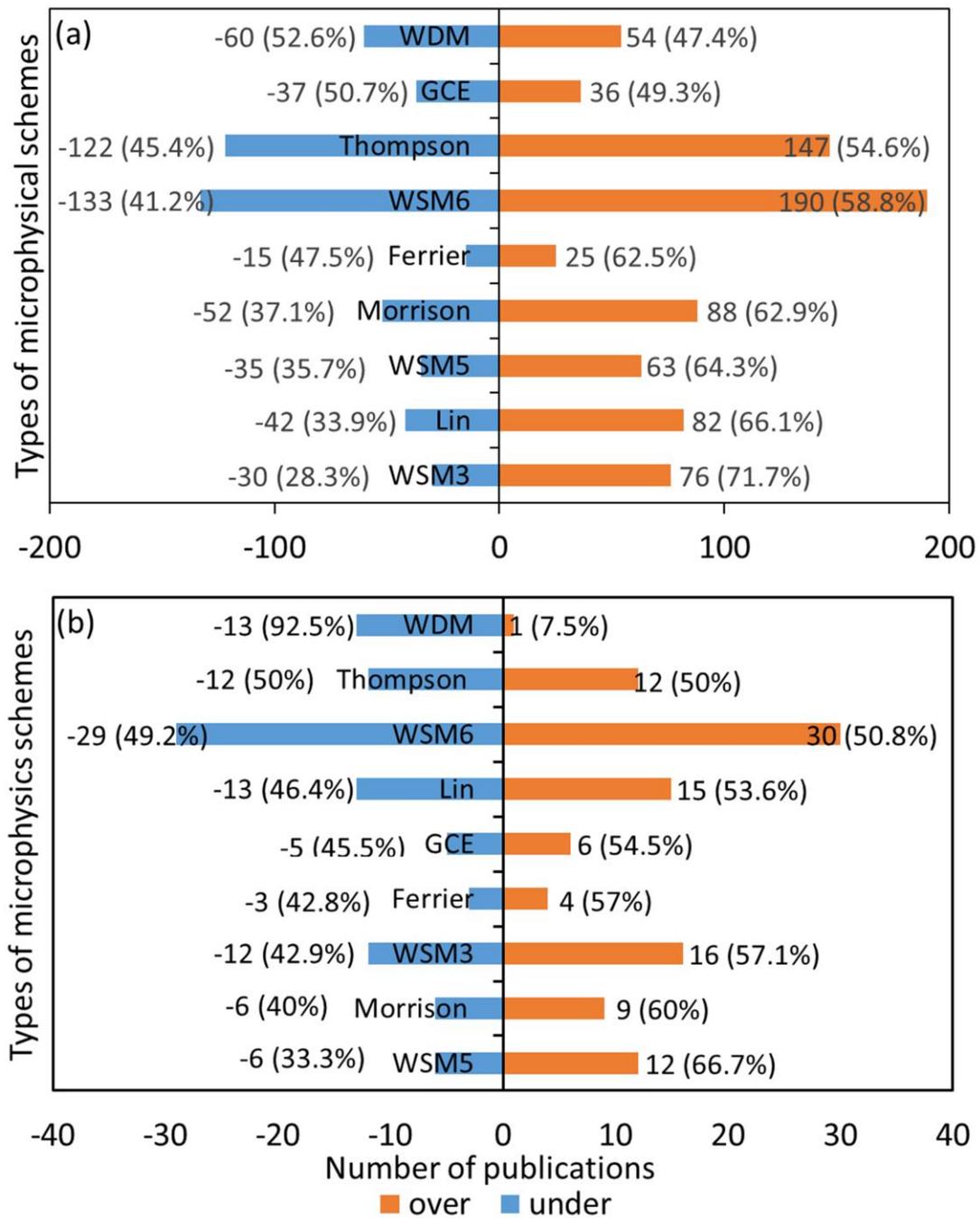

Figure 10. The number of publications considering (a) all configurations, (b) only the most popular configurations (RRTM/Dudhia, KF, YSU, Noah) displaying a definitive sign of precipitation error ranked from top to bottom relative to increasing percentage of overestimation. Blue bars represent the number of publications with underestimation, and orange bars represent the number of publications with overestimation.

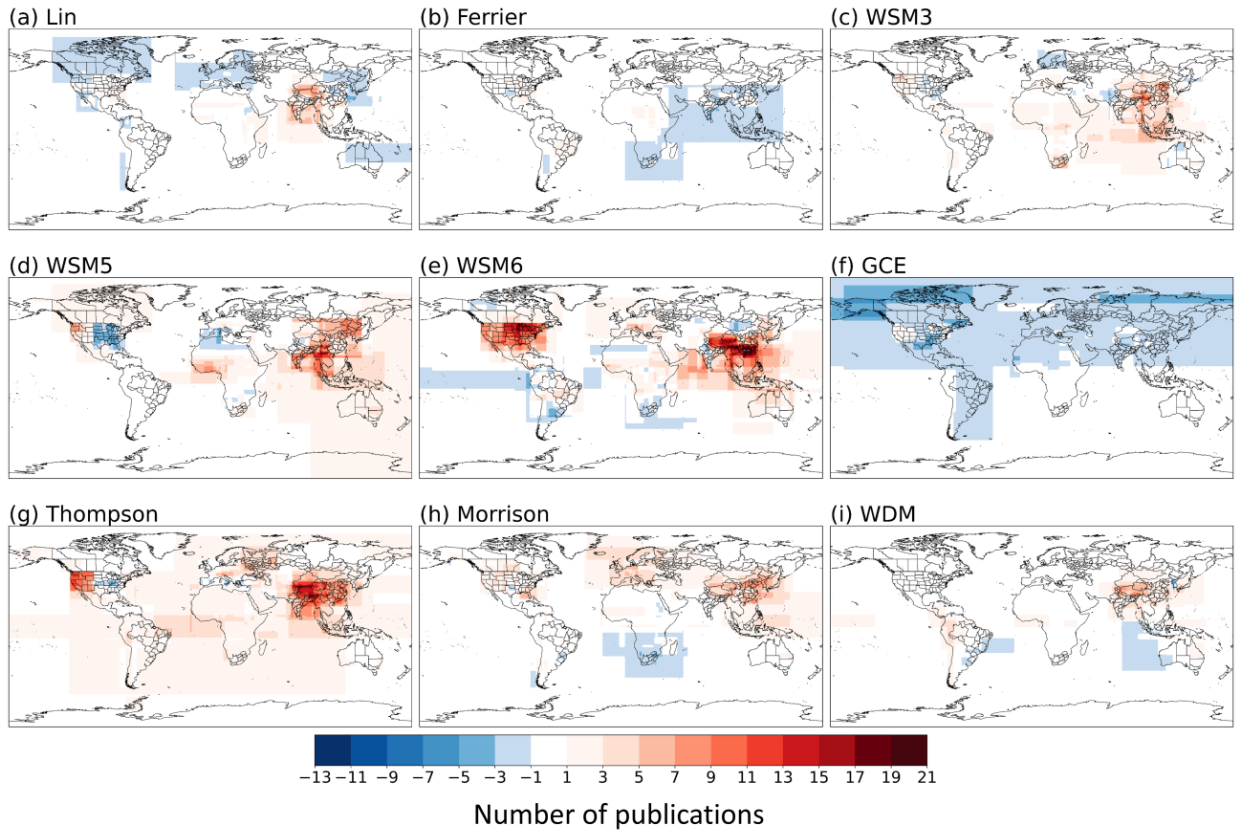

Figure 11. Global maps of differences between the number of publications that overestimated and underestimated precipitation using (a) Lin, (b) Ferrier, (c) GCE, (d) WSM3, (e) WSM5, (f) WSM6, (g) Thompson, (h) Morrison, and (i) WDM parameterizations. The first two rows are the one-moment microphysics parameterizations, and the third row is the double-moment parameterization. The orders of panels (a)–(f) and (g)–(h) are based on the year when the microphysics parameterization was added to the WRF model.

To quantify the performance of the nine microphysics parameterizations, we extracted RMSE and CC (correlation coefficients) of the accumulated precipitation between WRF simulations and observations (Fig. 12). The reason for using RMSE and CC is that, among the 300 publications selected for manual verification (section 2e), these two metrics were the most frequently used, with 45 and 57 publications, respectively. In contrast, the usage frequency of other metrics is much lower; for example, normalized root mean squared error (NRMSE) was mentioned in only three publications. In the present article, we chose RMSE with units of mm day$^{-1}$ because 69% (230 out of 334) of publications calculating RMSE used this unit. Among the 300 publications selected for manual verification (section 2e), the accuracy of questions 7 and 8 was as high as 90%, but some publications showed evaluation metric values in not only texts, but also tables and figures, which brought challenges for language models to extract information completely. Among these 300 publications, 23% (70 publications) provided RMSE and CC values, but GPT-4 Turbo could not retrieve complete data from 19 of them. Therefore, to build more comprehensive database, we manually extract RMSE and CC values from 575 publications based on the answers from GPT-4 Turbo.

Figure 12a shows the distribution of RMSEs for each microphysics parameterization. A small value of RMSE means a good prediction of precipitation. The Ferrier and WSM5 parameterizations performed better than other parameterizations with smaller median values of 2.19 and 2.99 mm day$^{-1}$, respectively, and smaller values of 1.5 times the interquartile ranges (i.e., the length of the whiskers) from near 0 to about 13 mm day$^{-1}$, indicating that these parameterizations tended to predict the accumulated precipitation closer to the observations than the other parameterizations. However, the Lin and Morrison parameterizations performed much worse, with larger median values (8.25 and 8.78 mm day$^{-1}$, respectively) and large values of 1.5 times the interquartile range (up to 37 mm day$^{-1}$). Although all distributions are skewed to higher values of RMSE, the degree of skewness varies. For example, WSM3 and WDM have comparable medians, but the 25th percentile for WDM is less than that for WSM3, indicating that WDM yields slightly better predictions in bulk. Also, some parameterizations have small interquartile ranges (e.g., Ferrier, WSM3, WSM5), whereas others have much larger interquartile ranges, albeit some have larger medians (e.g., Lin, Morrison) than others (e.g., WSM6, GCE, Thompson, WDM).

Figure 12b shows the distribution of CC for each microphysics parameterization. A value of CC approaching 1.0 means nearly perfect grid-point match between precipitation amounts from the model and the observations. The GCE parameterization performed better than other parameterizations with the larger medium value of 0.83 and a narrow distribution of 1.5 times the interquartile range of 0.5 and 1, indicating a relatively better performance. The WSM6 and Thompson parameterizations also performed better with median values of 0.76 and 0.78, respectively, following the GCE parameterization. The Lin parameterization has a value of the median of 0.72 in the middle range of all nine schemes, but a narrow interquartile range. However, the WSM5 and WDM parameterizations performed the worst of the nine parameterizations with medians of about 0.6 and the widest distributions of 1.5 times the interquartile range of 0 and 1. The distributions of 1.5 times the interquartile range for CC are less skewed than those for RMSE. In general, the Ferrier parameterizations performed best in RMSE, but the GCE parameterization performed best in CC. However, their sample sizes are small (39 and 68, respectively).

Looking at only those publications calculating both RMSEs and CCs, Fig. 13 shows scatterplots of RMSE and CC for each microphysics parameterization. For Ferrier, WSM6, GCE, and Thompson parameterizations, points tended to be concentrated in the upper left corner, representing good performances (Figs. 13b,e,f,g). However, for the remaining parameterizations, points were relatively scattered (Figs. 13a,c,d,h,i). For WSM3, WSM5, WSM6, Thompson, and Morrison parameterizations, many well-performing simulations with RMSEs were less than 1 mm day$^{-1}$ (Figs. 13c,d,e,g,h). Except the Ferrier and GCE parameterizations, the other seven parameterizations combined to produce 38 poor samples in which RMSE exceeded 100 mm day$^{-1}$ and CCs were negative (Figs. 13a,c,d,e,g,h). All of these points came from publications related to extreme weather systems, such as a hurricane in the United States using the Morrison parameterization (Sikder et al., 2019), a typhoon in China using the WSM5 parameterization (Fang et al., 2011), a heavy rainfall event in India using the Thompson parameterization (Rajeswari et al., 2021), a heavy rainfall event in South Korea using the WSM5 parameterization (Kwon & Hong, 2017), and a heavy rainfall event in China using the WSM6 parameterization (Ying et al., 2022). This finding indicates that models can face challenges in verifying simulations of extreme precipitation events.

Another cluster of points occurs in Figs. 13c,d,g,h where a line of points slopes steeply down just shy of 100 mm day$^{-1}$ RMSE. These points come from two studies: Jang and Hong (2016) for Fig. 13c and Chen et al. (2017) for Figs. 13d,g,h. At least for the points from Chen et al. (2017), these clusters of points in roughly the same locations regardless of microphysical parameterization used suggests that errors in the microphysics parameterization is unlikely to be the reason these points have relatively poor performance. How the points from Jang and Hong (2016) relate is unclear.

In Asia and North America, the data points appeared relatively scattered across all schemes (Fig. 13). In Europe, RMSE values clustered around 1 mm day$^{-1}$ with the WSM5 and Morrison schemes (Figs. 13d,h) but increased to approximately 10 mm day$^{-1}$ with the WSM3, WSM6, Thompson, and WDM schemes (Figs. 13c,e,h,i). Limited data were available for South America and Africa (Figs. 13); however, most of these data exhibited low RMSE and high correlation coefficient, indicating better model performance in these studies.

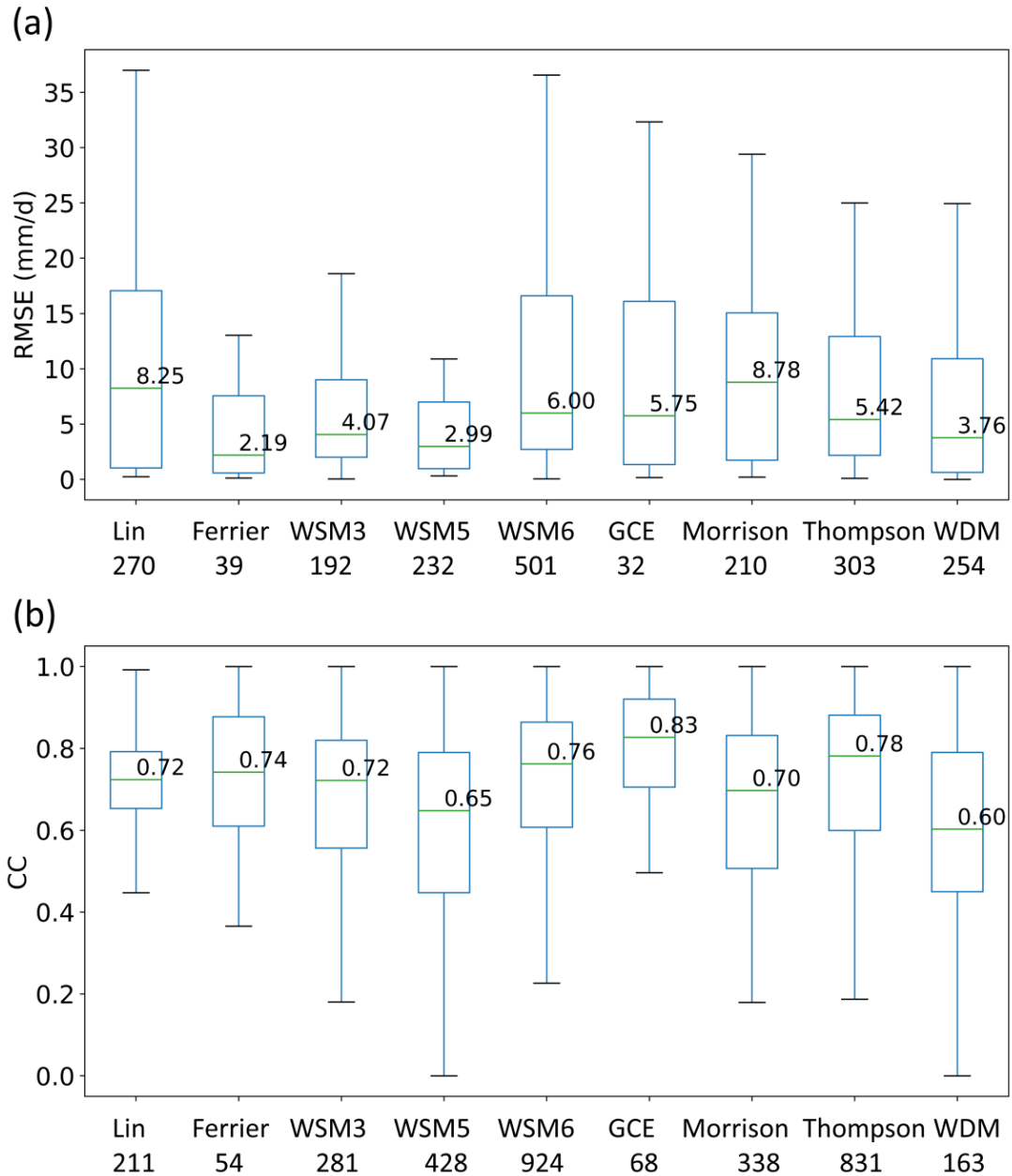

Figure 12. Boxplots of (a) RMSE (mm day⁻¹), and (b) CC of accumulated precipitation between simulations and observations among nine microphysics parameterizations. Numbers near the bars represent the medians, and numbers at the bottom represent the sample size for different microphysics parameterizations. The central box represents the 25th, 50th (median), and 75th percentiles, whereas the whiskers extend to a length of 1.5 times the interquartile range.

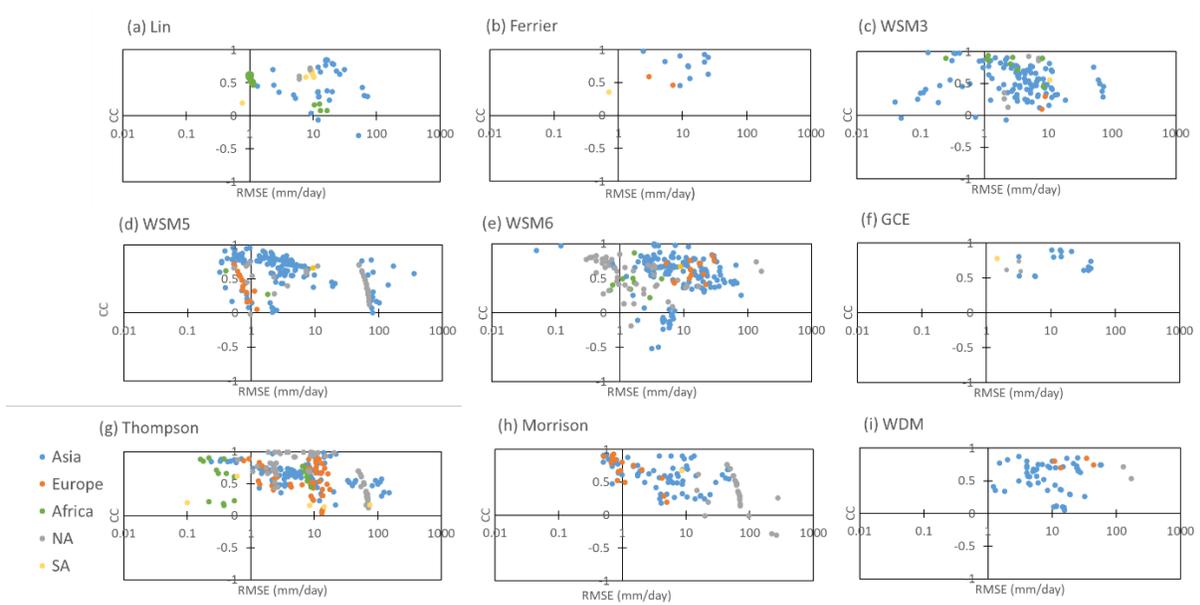

Figure 13. Scatterplots of RMSE and CC for the (a) Lin, (b) Ferrier, (c) WSM3, (d) WSM5, (e) WSM6, (f) GCE, (g) Thompson, (h) Morrison, and (i) WDM parameterizations. Colors represent different continents (Asia = blue, Europe = red, Africa = green, North America = grey, South America = yellow).

**6 Conclusions**

We constructed a quantitative review of existing studies on precipitation simulations with the Weather Research and Forecasting (WRF) model. We collected 2803 relevant publications from Web of Science and Scopus from 2003 to 2023. We conducted an analysis of the development of the WRF model in precipitation simulation, examining the number of publications, citation counts, author distribution, and their affiliated countries and research institutions. The results indicate that from 2003 to 2023, the number of publications on WRF precipitation simulation has increased. *Monthly Weather Review* is the journal with the highest mean citation counts per publication, while *Journal of Climate* has the highest median citation counts per publication. Additionally, the number of authors per publication has been increasing, with most authors affiliated with research institutions in the United States and China. This trend highlights the interest into this field and the substantial investment made by research institutions in precipitation simulation studies.

We found 2699 PDF files out of 2803 publications, then asked GPT-4 Turbo eight questions to extract information about model configurations and performance. To verify the performance of GPT-4 Turbo, we compared the answers from these eight questions for 300 publications using a manual method versus the GPT-4 Turbo method. Seven out of eight questions had an accuracy rate above 90%. Only one question about the performance of the WRF model had a comparatively lower accuracy at 86%. These results indicate that the information extracted by GPT-4 Turbo provides mostly reliable information for our study.

We conducted statistical analysis on the data produced by the GPT-4 Turbo and obtained the following main findings:

• Bulk microphysics parameterizations were more commonly used in the precipitation simulations (up to 95%) than bin microphysics parameterizations (5%). In bulk microphysics parameterizations, the WRF Single-Moment 6-class (WSM6) parameterization was most popular, with 776 publications adopting this parameterization to simulate precipitation. The second most popular parameterization was the Thompson parameterization (678 publications). These two parameterizations were more than twice as likely to be used in publications compared to the other microphysics parameterizations.

• Before 2020, the one-moment microphysics parameterizations were dominant. Among these parameterizations, the WSM6 parameterization was the most popular. However, the popularity of one-moment parameterizations decreased with year from 100% in 2004 to 51% in 2020, as the popularity of double-moment parameterizations grew. After 2020, double-moment microphysics parameterizations became the dominant parameterization (about 53%), the most popular of which was the Thompson parameterization.

• A new parameterization usually began to be used in precipitation simulations one or two years after the parameterization code was added to the WRF code because the time needed to conduct the research using a new scheme and then write a manuscript about it (and have it published).

• The geographical distribution of the number of relevant publications varies depending on the type of microphysics parameterizations. Specifically, the Lin, WRF Single-Moment 3-class and 5-class (WSM3 and WSM5), and WRF Double-Moment (WDM) parameterizations were commonly used in India and China, whereas the Ferrier, Thompson, and Morrison parameterizations were more commonly used in the United States.

• Regardless of the microphysics parameterization chosen, the most commonly used configuration was the Rapid Radiative Transfer Model (RRTM)/Dudhia long and short radiation, Kain–Fritsch (KF) cumulus, Yonsei University (YSU) planetary boundary layer, and Noah land surface parameterizations (500 publications using this configuration).

• Seven out of nine microphysics parameterizations tended to overestimate the precipitation. However, for the most popular configuration of each microphysics parameterization, the degree of overestimation was reduced.

• The Ferrier parameterizations performed best in RMSE (median of 2.19 mm day$^{-1}$), but the GCE parameterization performed best in CC (median of 0.83). However, their sample sizes are small (39 and 68, respectively).

• Some publications calculated large RMSEs over 100 mm day$^{-1}$ and negative CCs because these publications simulated extreme precipitation events such as a hurricane, typhoon, and heavy rainfall events. Therefore, models can face challenges in verifying simulations of extreme precipitation events.

In summary, quantitative analysis and review using GPT-4 Turbo provides a time- and labor-saving approach to systematically understand the usage and biases of microphysics parameterizations, as well as to provide a guide for the choice of microphysics parameterization and promote the development and accuracy of rainfall forecast simulation. However, we still face some challenges. For example, although our study highlights the tendency for overestimation in current microphysics parameterizations, we were unable to confirm the exact causes within this paper. This result underscores a critical need for future research. Our study is global, encompassing various climate zones and environmental contexts, making it more universally relevant and valuable for cross-regional and interdisciplinary readers. Additionally, our study has shown that it is possible to use LLM to answer scientific questions based on scientific literature. Looking forward, this kind of automated literature review methodology could be used for other kinds of difficult and persistent problems in atmospheric science. For example, it could be adapted to investigate biases in other types of model parameters and analyze the performance of different climate models on climate projections. The innovation of large language models to aid in conducting literature reviews is not only applicable to numerical weather or climate models but can also be extended to the large-scale literature analysis of other environmental topics.

**Acknowledgments**

This article is a contribution to Zhang's Ph.D. dissertation. We also thank the China Scholarship Council for supporting Shengnan Fu's studies in Manchester, which facilitated this project. Partial funding for Schultz was provided to the University of Manchester by the Natural Environment Research Council through Grants NE/N003918/1, NE/W000997/1, and NE/X018539/1. This work was also made possible by his time at the Aspen Center for Physics, which is supported by National Science Foundation grant PHY-2210452. CRediT statement: TZ: Conceptualization, Formal analysis, Investigation, Visualization, Writing – original draft; SF: Conceptualization, Formal analysis, Investigation, Visualization, Writing – original draft; DMS: Conceptualization, Supervision, Writing – review & editing; ZZ: Conceptualization, Supervision, Writing – review & editing.

**Open Research**

Web of Science is a subscription-based service, and access was obtained through the University of Manchester's subscription (https://www.webofscience.com/wos/woscc/basic-search). Scopus is freely available through the following archive: https://www.scopus.com/search/form.uri?display=basic#basic. Global geographic boundary data used in Figure 6(a) was provided by: Runfola, D. et al. (2020) geoBoundaries: A global database of political administrative boundaries. PLoS ONE 15(4): e0231866. https://doi.org/10.1371/journal.pone.0231866 [doi.org].